\documentclass{article}

\usepackage{arxiv}

\usepackage{amsmath}
\usepackage{multirow} 

\usepackage[utf8]{inputenc} 
\usepackage[T1]{fontenc}    
\usepackage{hyperref}       
\usepackage{url}            
\usepackage{booktabs}       
\usepackage{amsfonts}       
\usepackage{nicefrac}       
\usepackage{microtype}      
\usepackage{lipsum}
\usepackage{graphicx}
\graphicspath{ {./images/} }

\title{SafeLLM: Domain-Specific Safety Monitoring for Large Language Models: A Case Study of Offshore Wind Maintenance}

\author{
 Connor Walker \\
 University of Hull, Cottingham Road \\
 Hull HU6~7RX, UK \\
 AURA CDT, Hull, UK \\
 \texttt{C.Walker-2018@hull.ac.uk} \\
 \And
 Callum Rothon \\
 University of Hull, Cottingham Road \\
 Hull HU6~7RX, UK \\
 AURA CDT, Hull, UK \\
 \texttt{C.Rothon-2017@hull.ac.uk} \\
 \And
 Koorosh Aslansefat \\
 University of Hull, Cottingham Road \\
 Hull HU6~7RX, UK \\
 AURA CDT, Hull, UK \\
 \texttt{K.Aslansefat@hull.ac.uk} \\
 \And
 Yiannis Papadopoulos \\
 University of Hull, Cottingham Road \\
 Hull HU6~7RX, UK \\
 AURA CDT, Hull, UK \\
 \texttt{Y.I.Papadopoulos@hull.ac.uk} \\
 \And
 Nina Dethlefs \\
 University of Hull, Cottingham Road \\
 Hull HU6~7RX, UK \\
 AURA CDT, Hull, UK \\
 \texttt{N.Dethlefs@hull.ac.uk} \\
}

\begin{document}
\maketitle

\begin{abstract}
The Offshore Wind (OSW) industry is experiencing significant expansion, resulting in increased Operations \& Maintenance (O\&M) costs. Intelligent alarm systems offer the prospect of swift detection of component failures and process anomalies, enabling timely and precise interventions that could yield reductions in resource expenditure, as well as scheduled and unscheduled downtime. This paper introduces an innovative approach to tackle this challenge by capitalising on Large Language Models (LLMs). We present a specialised conversational agent that incorporates statistical techniques to calculate distances between sentences for the detection and filtering of hallucinations and unsafe output. This potentially enables improved interpretation of alarm sequences and the generation of safer repair action recommendations by the agent. Preliminary findings are presented with the approach applied to ChatGPT-4 generated test sentences. The limitation of using ChatGPT-4 and the potential for enhancement of this agent through re-training with specialised OSW datasets are discussed.
\end{abstract}

\section{Introduction}
As global demand for renewable energy increases in line with sustainability goals, the Offshore Wind (OSW) sector is poised to grow rapidly, expected to supply 25 \% of  electricity globally by 2050, circa 1,150 GW \cite{smith2023targets}. Nearly \nicefrac{1}{3} of the Levelised Cost of Electricity (LCoE) is from Operations and Maintenance (O\&M), including inspections, routine servicing, and repairs \cite{BVGGuide}. 

A significant shift towards Condition Based Maintenance (CBM) is driven by the increased complexity of turbines and distance from shore of sites. Advances in embedded sensors within turbines have provided access to detailed monitoring in real-time using Supervisory Control and Data Acquisition (SCADA) systems, facilitating remote diagnostics. 

Operational wind farms generate large datasets, constituting significant volumes of nuisance, false, and "chattering" alarms \cite{Wei2023clustering}. As an example, Teesside wind farm has experienced up to 500 alarms in a 24-hour period, correlating to an alarm every three minutes \cite{walker2022deep}. Nuisance alarms impede the accurate diagnosis of faults, causing delays to required maintenance. 

In this context, Large Language Models (LLMs) can potentially offer a useful aid to maintenance, though their application in safety-critical contexts requires reliability and trustworthiness as a precondition. Despite growing acceptance professionally, concerns regarding hallucinations \cite{huang2023survey} and unsafe responses \cite{inan2023llama} persist. While safety measures against risky inputs and outputs have been suggested \cite{meta2023guide}, susceptibility to circumvention remains a concern \cite{rando2022redteaming}. 

In Section 2, we discuss the existing literature for LLMs, focusing on safety and hallucination detection within the models. Section 3 presents the research questions. Section 4 details the proposed methodology for SafeLLM, with Section 5 presenting and discussing preliminary results. Section 6 suggests scope for future work and final conclusions follow in Section 7.
\section{Literature Review}
\subsection{Wind Turbine Alarms}
Currently, O\&M planning in the OSW industry is based around alarms from SCADA data, highlighting faults as they occur, complemented by routine inspections, including rope-access and Unmanned Aerial Vehicle (UAV) inspections \cite{BVGGuide}. Recent work in the OSW sector has focused on the diagnosis of faults from alarm data, the prediction of subsequent alarms based on sequences of previous alarms \cite{walker2022deep, wei2023embedding, Wei2023clustering}, reduction of chattering alarms leading to alarm overload \cite{Gonzalez_2016}, and decision support for O\&M staff \cite{Chatterjee20decisionsupport, Chatterjee22automatedQA}. Research on alarm prediction has made use of word embeddings and word2vec \cite{mikolov2013efficient} from Natural Language Processing (NLP) , treating alarms in a sequence as analogous to words in a sentence, allowing subsequent alarms to be predicted \cite{wei2023embedding, Wei2023clustering}. 

\cite{walker2022deep} proposes a system based on Long Short-Term Memory (LSTM) and Bidirectional LSTMs (BiLSTMs) for prediction of repair actions from sequences of alarms in the OSW domain. This departs from previous works focusing on predicting subsequent faults as opposed to the required action. 

\subsection{Large Language Models} 
In recent years, progress on LLMs has been rapid, leading to wide interest in their adoption for a range of tasks.  \cite{zhao2023survey} reviews recent progress on LLMs, including pre-training, adaptation tuning, utilisation, and capacity evaluation. 

Open AI’s Chat GPT \cite{openai2023gpt4} is perhaps the most widely used LLM-based chatbot, based on  GPT-3 \cite{brown2020language}.  GPT-4 \cite{openai2023gpt4}, the most recent development in the GPT family,  has shown high performance in diverse applications. Some sources have reported human-level capabilities, although the precise definition of “human-level” is debatable and open to interpretation.

\cite{Lukens23evaluating} investigates how ChatGPT may be applied to the automation of maintenance planning, identifying a range of criteria for assessment. The performance of ChatGPT is examined, and the risks and business case for adoption are discussed. \cite{li2023chatgptlike} reviews how recent developments in LLMs are impacting the fields of prognostics and health monitoring, and provides recommendations for future development of models in these fields, including interpretability and security.

\cite{touvron2023llama} presents LLaMA, a set of open-source LLMs with a range of model sizes, trained on publicly available data. Models included in LLaMA are competitive with the state of the art, so are considered a valuable resource to the research community, and offer more flexibility and transparency than leading models.

Methodologies have been presented to assess the quality of LLM-generated content on a range of measures. TRUE \cite{honovich2022true} is a standardised methodology for the assessment of factual consistency using a diverse range of sources, which aims to provide clear and actionable quality measures for generated content. GPTScore \cite{fu2023gptscore} is a framework that uses generative pre-trained models to score generated texts against 22 aspects, including factuality, accuracy, and consistency. G-Eval \cite{liu2023geval} proposes the use of LLMs and Chain of Thoughts (CoTs) for the evaluation of generated texts, showing high performance when using GPT-4 as the backbone on a summarisation task, achieving high correlation with humans. 

\subsection{Hallucination Detection}
A significant issue encountered with LLMs is that they can make non-factual and inaccurate statements, known as hallucinations.  In the GPT-4 technical report \cite{openai2023gpt4}, hallucinations are identified as a key issue with current LLMs, becoming more pressing as users build trust with LLMs. In a benchmarking study comparing LLMs to humans in question answering tasks, \cite{lin2022truthfulqa} reports that the best-performing models at the time were truthful 58 \% of the time, while humans were truthful 94 \% of the time, with models mimicking common human misconceptions.

\cite{huang2023survey} presents a survey of hallucination in LLMs, including a taxonomy of hallucinations and identification of causes. Hallucinations can be broadly grouped as intrinsic and extrinsic hallucinations \cite{maynez2020summarization}. Intrinsic (or closed-domain) hallucinations arise from the source data, while extrinsic (or open-domain) hallucinations ignore the source data. 

Detection of hallucinations has been a field of rapid development, in tandem with the rise of LLMs. One approach to avoiding issues arising from hallucinations is to take steps to prevent them from occurring. In a benchmarking study using TruthfulQA,  \cite{lin2022truthfulqa}  determined that the largest models were the least truthful overall, and that fine-tuning smaller models for specific tasks would be beneficial in terms of truthfulness. \cite{rateike2023weakly} proposes a method for the detection of hallucinations in LLM activations from pre-trained models, scanning the internal states using an anomaly-free reference dataset.

Another approach is to detect hallucinations in generated content from the LLM, and then to remove or correct the offending parts of the output. SelfCheckGPT \cite{manakul2023selfcheckgpt} is a hallucination detection algorithm which is capable of fact-checking outputs from LLMs without external resources by comparing stochastically sampled responses. The offline approach proposed is effective in many applications but can fail if hallucinations are present in all sampled sentences.  Chatprotect \cite{mündler2023selfcontradictory} aims to detect self-contradictory hallucinations in generated content from black-box LLMs, achieving 80 \% F1 score for detection of self-contradictory hallucinations on ChatGPT. The detected hallucinations are then refined by a mitigation algorithm to preserve fluency in the generated text. Chainpoll \cite{friel2023chainpoll} is a hallucination detection methodology that uses the novel metrics of correctness and adherence to detect hallucinations in LLM outputs. Chainpoll is tested against other methods using, RealHall, a collection of benchmark datasets also presented by this work.

\subsection{LLMs and Safety}
As Machine Learning (ML) based systems are being deployed in safety-critical applications with limited human oversight, it is vital that their dependability and robustness can be guaranteed. \cite{bianchi2024safetytuned} finds that most popular instruction following LLMs are susceptible to making unsafe responses when prompted. \cite{dong2024attacks} presents a comprehensive survey on LLM conversation safety, considering Attacks, Defences, and evaluation methods, and presents a taxonomy of studies to date on LLM conversation safety. The defences identified are categorised into three groups: input/output filters, inference guidance, and LLM safety alignment.  This survey also identifies a range of datasets for LLM safety evaluation, which mostly focus on toxicity, discrimination, privacy, and misinformation. No specific datasets for unsafe instructions which may cause injury were identified in this study.

In 2022, \cite{levy2022safetext} presents SafeText, a benchmarking dataset for commonsense physical safety in LLM generated instructions containing 367 scenarios with safe/unsafe advice pairs. The benchmarking study finds that many LLMs can generate unsafe text, and struggle to reject responses that may lead to physical harm. \cite{hawkins2021guidance} presents Assurance of ML in Autonomous Systems (AMLAS) and follows this with Safety Assurance of Autonomous Systems in Complex Environments (SACE) \cite{hawkins2022guidance}. AMLAS contains safety case patterns and processes for integrating safety into the development of ML components and justifying the acceptable safety of said components. SACE extends this work to fully autonomous systems. 

Rando J. et al. \cite{rando2022redteaming} finds that safety filters on stable diffusion image-generation models can be relatively easily bypassed and argues for a community-based approach to safety measures in generative Artificial Intelligence (AI). Their findings demonstrate the need for robust safety measures which are resistant to adversarial attacks. \cite{inan2023llama} presents LLaMA Guard, a safeguarding model built around models in LLaMA using a safety risk taxonomy to classify prompts and responses. This work presents a taxonomy of safety risks that may arise when interacting with an AI agent, including violence and hate speech, sexual content, and criminal planning. LLaMA Guard takes this taxonomy as input and classifies user inputs (prompts) and agent outputs as encouraged (safe) and discouraged (unsafe) using a single model. By using a different taxonomy, the model can be fine-tuned using zero-shot and few-shot methods.  

\section{Research Questions}
In 2020, discussions between experts on LLMs, including representatives from OpenAI, and the Stanford Institute for Human-Centered AI, identified several research questions regarding the future of LLMs. One such research question was -- \textbf{What can academia do to best position itself to develop guardrails for the industrial development of such models \cite{tamkin2021understanding}? } To address this, and the aforementioned issues with  LLMs in the context of OSW Maintenance, in this paper,  we propose a novel approach called SafeLLM which has the following features:
\begin{itemize}
    \item Statistical methodology for unsafe response prevention in an embedded safety layer.
    \item  Hallucination detection to identify incorrect or irrelevant responses.
\end{itemize}

We define essential concepts, describe our methodology, and present preliminary results from an OSW application. The discussion offers insights into our findings, and the concluding section outlines future research directions, including the potential application of Reinforcement Learning (RL).

\section{Methodology}
Existing model frameworks predominantly focus on the application of Cosine similarity to generate similarity scores of sentence embeddings, obtained from both the user input and generated output. However, if the response generated is previously unseen, there is a potential risk it has started to hallucinate from the outset.

We propose the use of additional Empirical Cumulative Distribution Function (ECDF) statistical distance measures, namely Wasserstein distance, also known as Earth Mover's Distance (EMD), to calculate sentence similarity. To benchmark our results against existing methodologies, each sentence is tested on both measures: Cosine similarity and EMD.

\subsection{Cosine Similarity}
Cosine Similarity is a distance metric used to measure the similarity of two vectors by calculating the cosine of the angle between them. This methodology disregards magnitudinal differences and scale. Simply defined, it is the dot product of each vector divided by their respective magnitudes: $ \cos(\theta) = \frac{A \cdot B}{\parallel A \parallel \parallel B \parallel} $, where $A$ and $B$ are the input vectors, or sentence embeddings, of the two sentences being compared.
\subsection{Wasserstein (EM) Distance}
EMD is a measure for comparison of probability distributions, proven effective in a diverse range of applications  \cite{Panaretos_2019}, including ML and fault detection. Based on the Optimal Transport Theory (OTT), the measure is considered the most efficient method of moving a mass distribution into another. OTT is defined in the practical formulation: 
\begin{center}
    \begin{math}
    T:X \rightarrow Y transports \ \mu \in \ P(X) \ to \ \nu \in P(Y)
\end{math}
\end{center} Cost can be determined by transporting one unit from $x \in X \ to \ y \in Y$ giving a further definition of OTT: \begin{center}
\begin{math}
    \inf_T \int || x - T(x) || \ p dP(x)
\end{math}\end{center}
Building on this theory, EMD considers the movement of distributions to multiple locations, defined as: \begin{center}\begin{math}
    W_{p}(P, Q) = \bigl( \inf_{J \in j(P,Q)} \int ||x - y|| \ pdJ(x, y) \bigl) 
\end{math}\end{center}
Where, $j(P,Q)$ denotes all joint distributions $J$ for $(X,Y)$ with marginals $P$ and $Q$. \\Calculating $P$ and $Q$ as the CDF of $p$ and $q$ therefore allows the common simplification of EMD: \begin{center}
 \begin{math}
    W_{p}(p,q) = \int_{-\infty}^{+\infty}|P - Q|
\end{math}\end{center}

\subsection{Data Collection \& Pre-Processing}
Confidentiality of data within the OSW sector provides limited access to maintenance records with valuable insight for training. Notes such as "WTG1A HV MAINTENANCE" cause difficulties in the comprehension and implementation of the suggested actions. As such, for testing feasibility of EMD as a statistical safety measure, we employ ChatGPT4 to suggest sentences both unsafe and safe, creating an "Unsafe Concepts Dictionary". It should be noted that safe categorisation of a sentence is therefore not verified beyond the existing safe training of ChatGPT, giving scope for future verification of more comprehensive data sources.

In the sentences' raw form, it is not possible to measure similarity using the aforementioned methods. Therefore, it is required that each sentence is embedded; a method used to encode sentences into vector form. In the work presented within this paper, we use the Universal Sentence Encoder (USE) transformer-based model. This model takes a string input and provides a fixed dimensional vector output. As with LLMs, the encoding model is pre-trained, meaning comparative analysis of other encoder models will be essential in finding the best suited to OSW application. 

\subsection{Training}
\begin{figure}[h!]
    \includegraphics[width=0.99\linewidth]{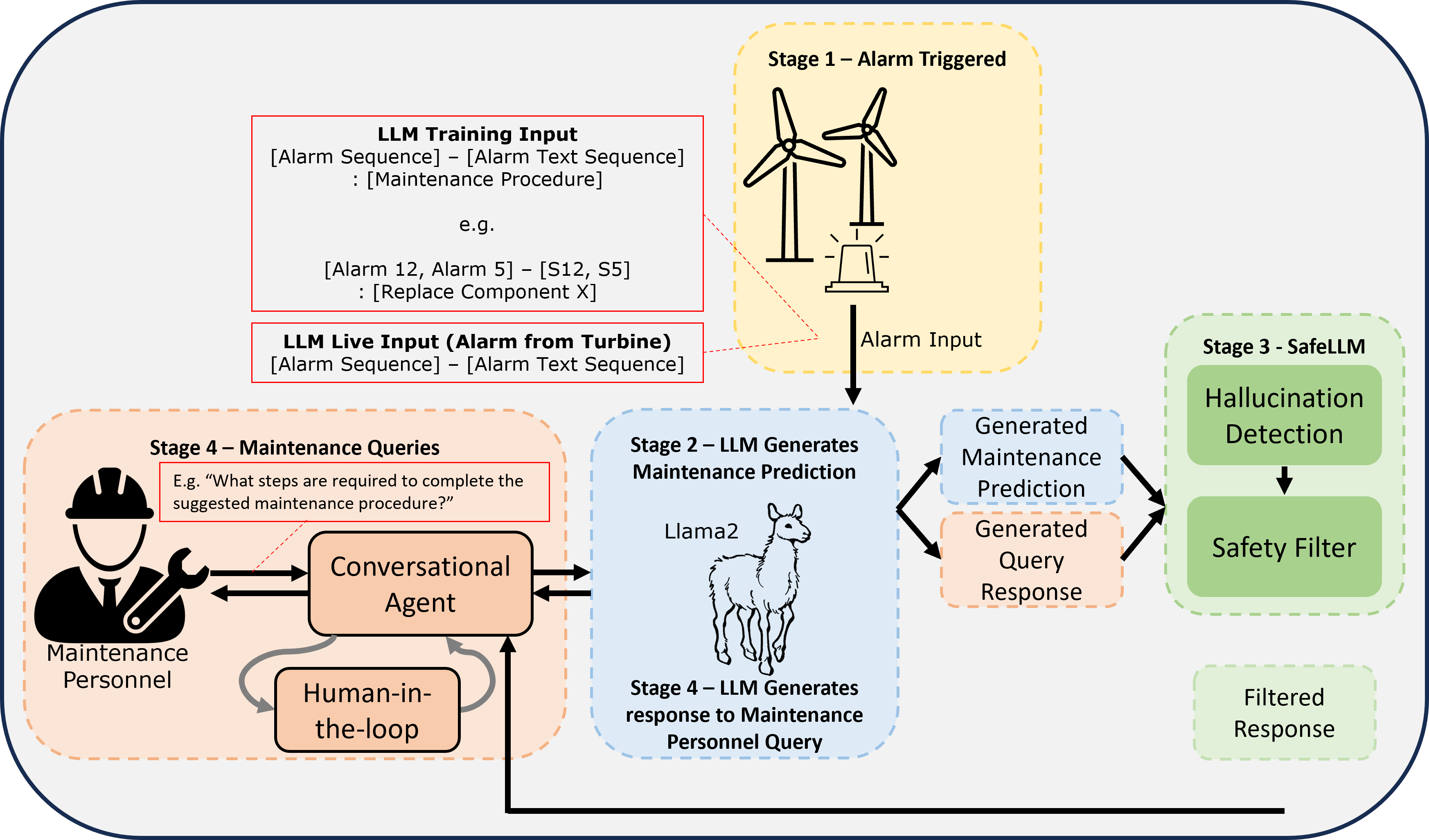}
    \label{trainingfig}
    \caption{Overview of the proposed SafeLLM approach -- training procedures}
\end{figure}

Given an input of an alarm, or sequences of alarms within a timeframe, the LLM should predict and recommend a maintenance procedure. Therefore, the training data is a concatenation of this form; the known maintenance procedure paired with correlated alarm sequences. A knowledge graph is also proposed, intended to support the ongoing training of an LLM as the complexity of the data increases with time. Once trained, generated responses are analysed through the feature layers of SafeLLM; Hallucination Detection, and Safety, which are discussed in upcoming sections. Once a prediction is presented, it is then intended that the maintenance personnel can interact with the LLM as a training aid to break down the maintenance action(s) where there is ambiguity or lack of knowledge. Figure \ref{trainingfig} shows an overall view of the proposed framework, including the training process.

Whilst Llama2 is shown as the LLM of choice (see stage 2), other models could be used. With access to real-world OSW data in the future, various models will be trained and fine-tuned to select the best performing for this application. Currently ChatGPT is utilised to generate test data.

\subsection{Safety Filter}
\begin{figure}[h!]
    \includegraphics[width=0.99\linewidth]{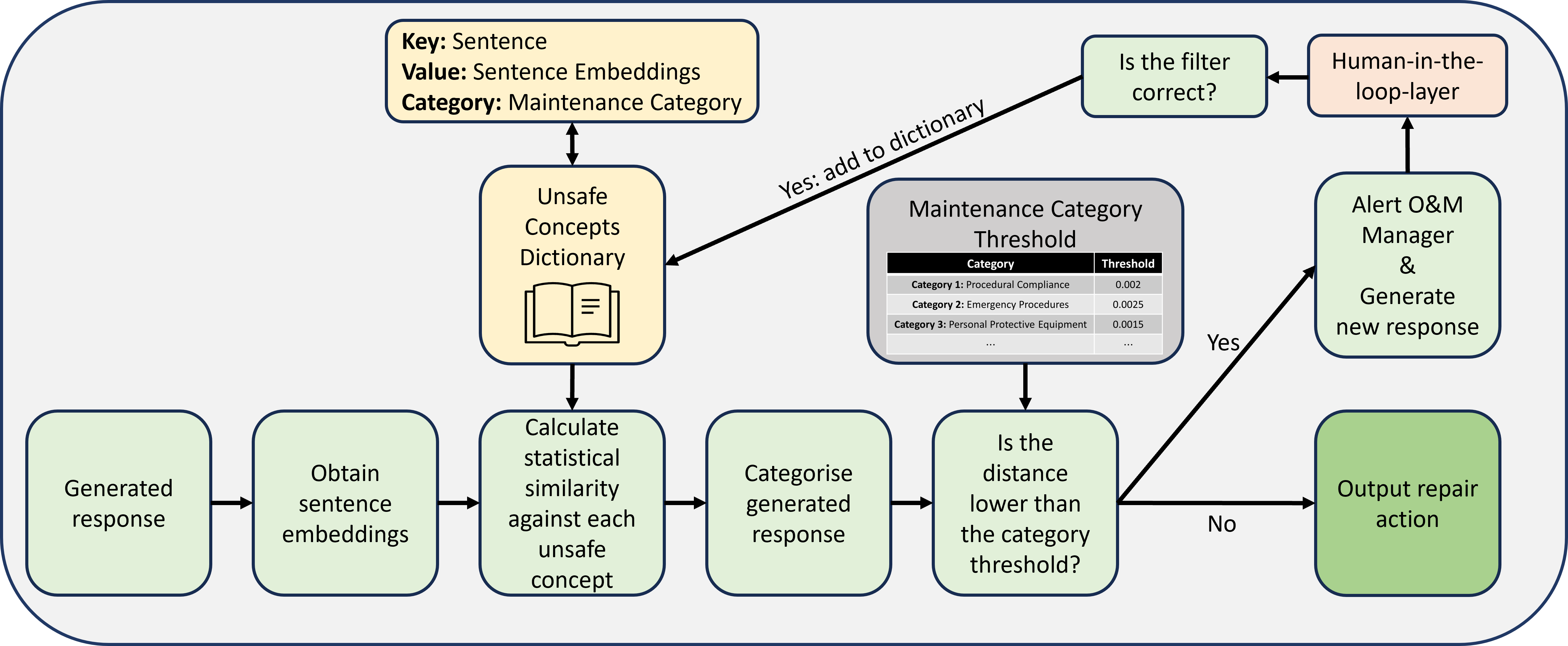}
    \label{safetyfilter}
\caption{Overview of the proposed SafeLLM approach -- and filtering unsafe responses using EMD}
\end{figure}

The SafeLLM framework obtains the generated responses' embeddings to calculate EMD against a pre-defined dictionary of unsafe concepts. Using the labelled category, we can categorise the response before comparing it to a fine-tuned threshold. For EMD, the lower the value, the higher the similarity. If the distance is lower than the threshold, we determine it unsafe and alert the O\&M manager. Where an unsafe response is detected, and verified as a correct identification, this is then added to the dictionary for future monitoring. 
Figure \ref{safetyfilter} suggests an overview of the framework. 

\subsection{Hallucination Detection}
Hallucinations occur where responses are generated from unseen inputs, allowing the model to respond without confidence of knowledge gained from training data. We look to address the issue from a view that given an input, the model should be capable of generating consistent outputs with high similarity scores; failing to achieve this suggests a hallucination has occurred. 

The LLM is proposed to generate $N$ responses for every input, where $N$ can be fine-tuned to optimise computational efficiency of the conversational agent whilst maintaining high-level accuracy. Each responses' similarity is then compared using EMD to identify hallucination occurrence. 

\begin{figure}[h]
\centering
\includegraphics[width=0.75\textwidth]{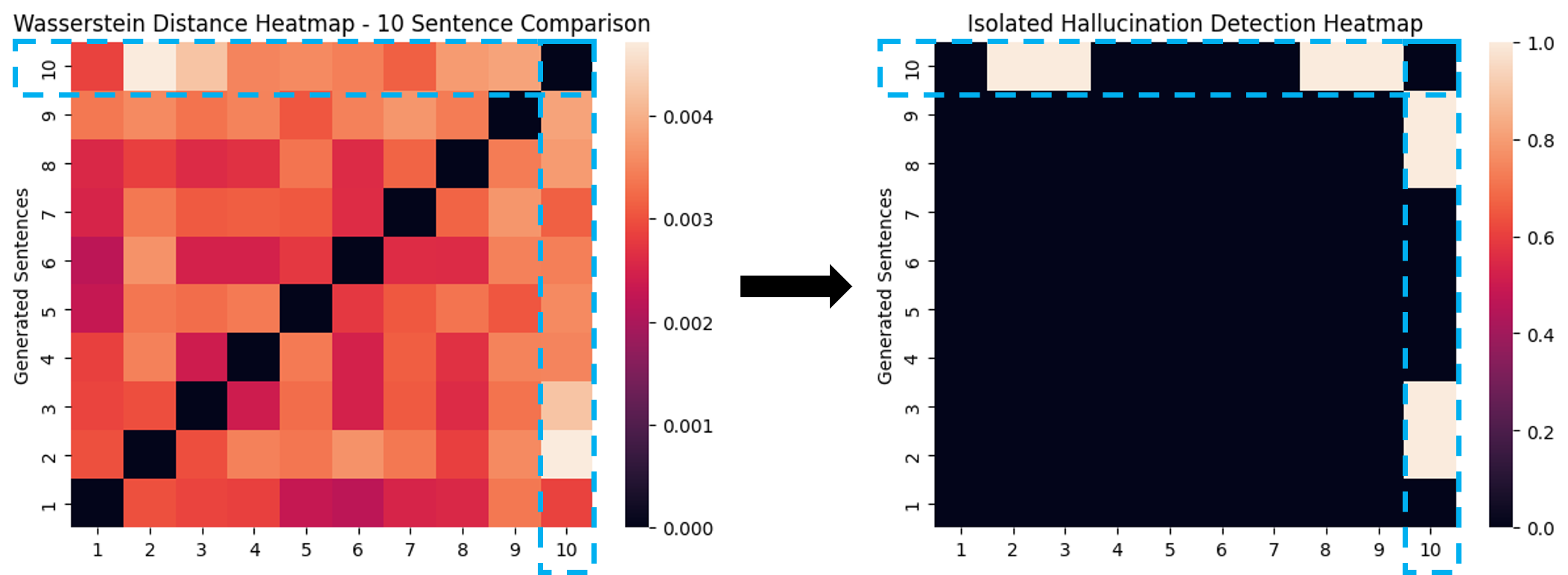}
\caption{ \label{hallucexample}10 generated sentences; nine acceptable, one hallucinated }
\end{figure}

Using a fine-tuned variance threshold, we are able to isolate sentences detected as a hallucination. In the example below, Figure \ref{hallucexample}, sentence 10 has hallucinated; sentences 1-9 are within an acceptable variance range. This threshold, being a hyper-parameter of SafeLLM, will be further fine-tuned dependant on the application. For this example, the limiting threshold is 0.0042, with an occurrence threshold of 40 \%; the limiting threshold being the maximum acceptable variance of sentence similarity, and the occurrence threshold being the minimum number of sentences that have to be above the variance for it to be identified as a hallucination.

\subsubsection{Hallucination Detection in Large Language Models}
\begin{itemize}
    \item[] \textbf{Step 1 -} Deviation Matrix Construction:
    \item[] Let \( H = \{h_1, h_2, \ldots, h_n\} \) be the set of hypotheses generated by the LLM for a series of inputs. Construct the deviation matrix \( R \) where each element \( r_{ij} \) is the deviation of hypothesis \( h_i \) from a factual baseline \( F \): \begin{math}
        r_{ij} = \text{Deviation}(h_i, f_j)
    \end{math}
\end{itemize}
\begin{itemize}
    \item[] \textbf{Step 2 -} Fidelity Constant Matrix, \(Fmatrix\):
    \item[] For each hypothesis \( h_i \), let \( f_i \) be the corresponding ground truth. Calculate the fidelity constant for each hypothesis: \begin{math}
        F_i = \frac{1}{1 + \text{Deviation}(h_i, f_i)}
    \end{math}
\end{itemize}
\begin{itemize}
    \item[] \textbf{Step 3 -} Consistency Calculation:
    \item[] Determine a threshold \( \theta \) for significant deviations. Calculate the consistency of \( R \) and \(Fmatrix\): \begin{math}
        C_R = \frac{\text{Number of elements in } R \text{ that are } < \theta}{n^2}
    \end{math}, \begin{math}
        C_F = \frac{\text{Number of elements in } Fmatrix \text{ that are } > 1 - \theta}{n}
    \end{math}
\end{itemize}
\begin{itemize}
    \item[] \textbf{Step 4 -} Combined Metric:
    \item[]  Create a combined metric \( M \) using weighted averages of the consistency values from \( R \) and \(Fmatrix\), with weights \( w_R \) and \( w_F \), respectively:
    \begin{math}
        M = w_R \cdot C_R + w_F \cdot C_F
    \end{math}
\end{itemize}
\begin{itemize}
    \item[] A higher value of \( M \) indicates better factual alignment and lower hallucination frequency, while a lower value suggests either significant hallucination, poor model grounding, or both.
\end{itemize}

\section{Results and Discussion}
This work reports on results using input data in the form of sentences generated by ChatGPT-4. Data is available from EDF's Teesside Windfarm, but  lacks comprehensible repair actions. Example maintenance actions include "Work in A4" and "Back Up Battery Error", with no clear indication of tangible repair actions. We are currently working with engineers to augment this data with maintenance actions associated with each alarm type, allowing the development of OSW-specific data in the future. These would in turn help to successfully train and test the LLM for prediction accuracy.
In the absence of such data, the accuracy of the proposed SafeLLM methodology is determined by comparing the 'safe' and 'unsafe' categorisation to that of ChatGPT's categorisation during generation. Examples of the generated sentences include, unsafe; \textit{"No fall protection measures should be required as the gearbox is away from hatches."}, and safe; \textit{"PPE is mandatory for all aspects of repair tasks."}.

Table \ref{comp_both_tbl} presents a comparison of Cosine similarity against EMD for each category. The accuracy is determined by the number of sentences correctly identified as safe and unsafe. During testing, each category threshold was incremented in stages, such as 0.005, between 0 and 1 to find the best-performing threshold, with only the highest accuracies for each being recorded.
\begin{table}[h]
\centering
\caption{\label{comp_both_tbl}Accuracy comparison of Cosine similarity (CS) \& EMD}
\begin{tabular}{@{}cccccccccccc@{}}
\toprule
\multicolumn{2}{c}{\multirow{2}{*}{}} & \multicolumn{10}{c}{\textbf{Category}} \\
\multicolumn{2}{c}{} & 1 & 2 & 3 & 4 & 5 & 6 & 7 & 8 & 9 & 10 \\ \midrule
\multirow{2}{*}{\textbf{\begin{tabular}[c]{@{}c@{}}Accuracy\\ (\%)\end{tabular}}} & CS & 87.5 & 92.5 & 67.5 & 92.5 & 67.5 & 62.5 & 67.5 & 70.0 & 92.5 & 82.5 \\
 & EMD & 50.0 & 75.0 & 52.5 & 72.5 & 62.5 & 70.0 & 75.0 & 60.0 & 72.5 & 85.0 \\ \bottomrule
\end{tabular}
\end{table}\\
Although Cosine similarity is shown to have higher accuracy in 7 of the 10 categories, it can be seen that EMD is within a comparable range for most categories. EMD and Cosine similarity show a similar pattern. With the exception of category 1 both methods achieve accuracy that is either higher than 72\% or lower than this in each category. It should be noted that the thresholds have had limited fine-tuning to show potential with the distance measure. We showed in our previous research that how fine tuning the threshold can be performed and how it can improve the outcome \cite{farhad2022keep}.

\begin{figure}[!ht]
\centering
\includegraphics[width=1\textwidth]{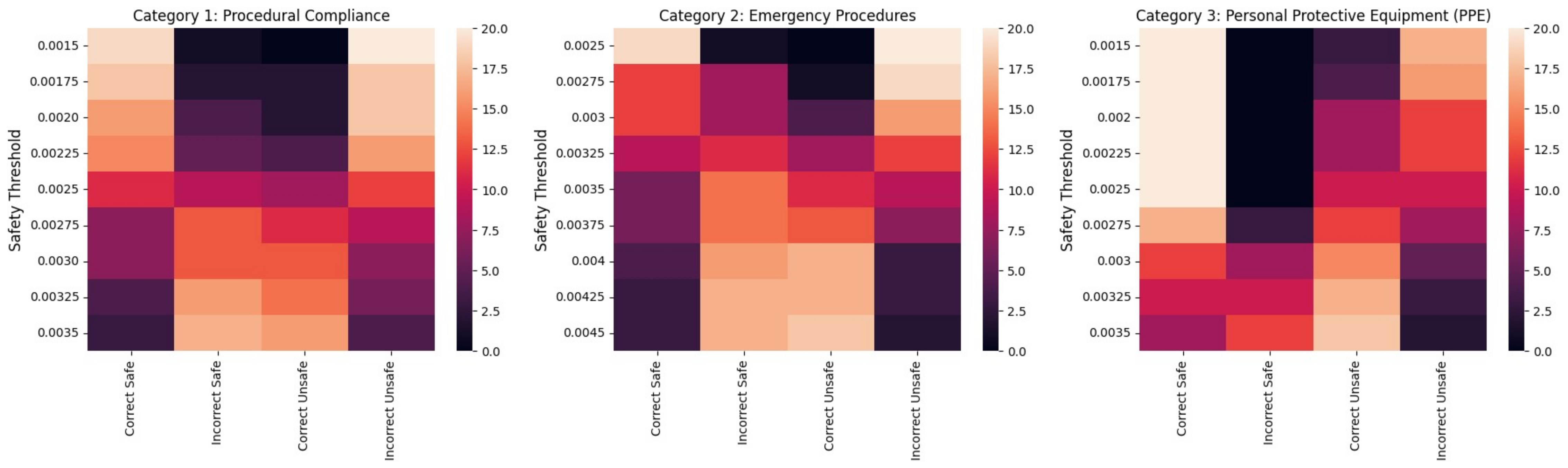}
\caption{\label{heatmap}Confusion matrices for Categories 1-3 at varying thresholds}
\end{figure}

\begin{figure}[!ht]
    \centering
    \includegraphics[width=0.9\textwidth]{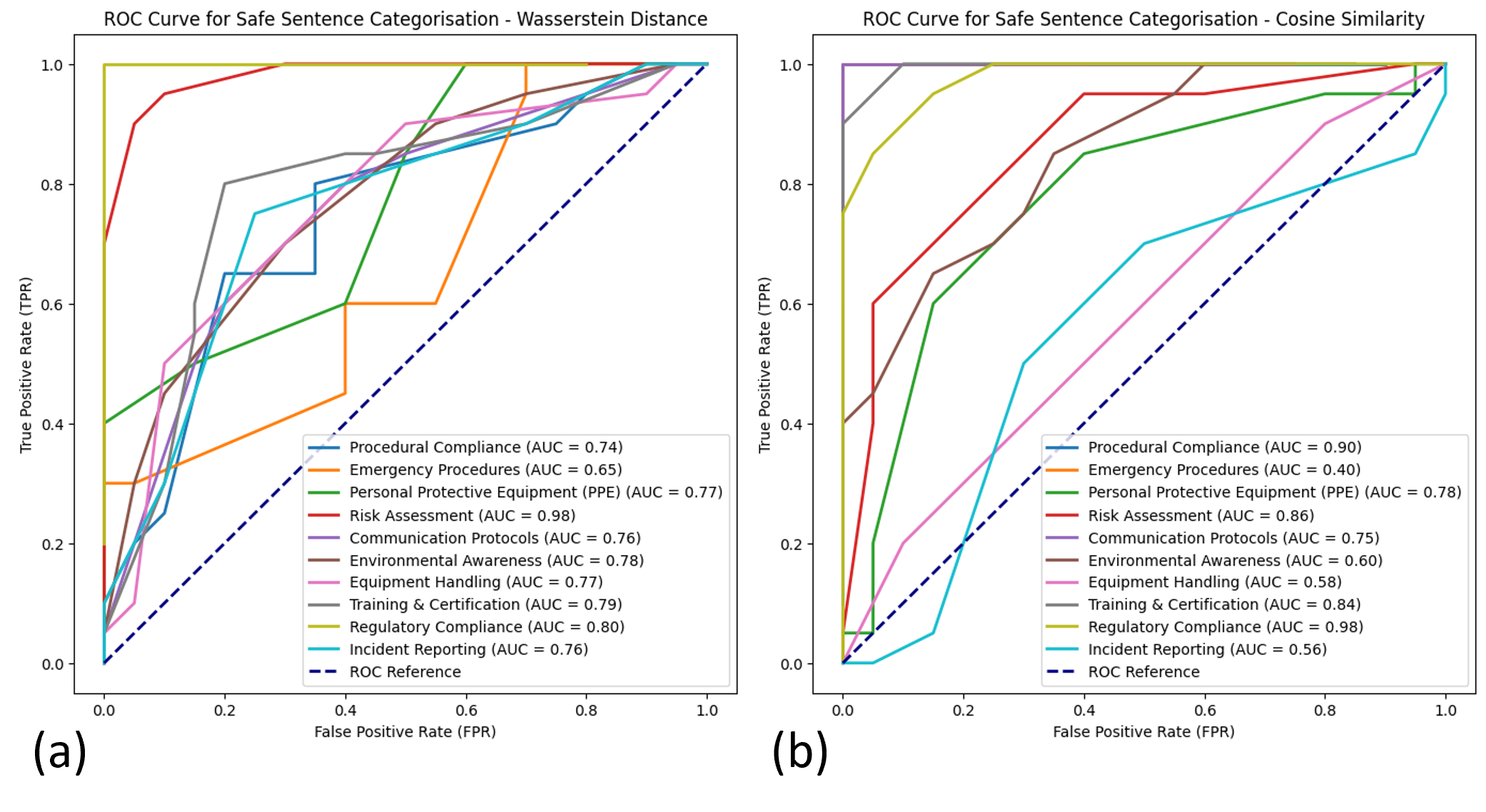}
    \caption{\label{fig_roc_curves}ROC Curves for Safe Sentences: (a)Wasserstein distance; (b) Cosine similarity}
\end{figure}

Figure \ref{heatmap} shows accuracy of 3 categories as a confusion matrix, showing the number of both correctly and incorrectly identified unsafe \& safe sentences at varying distance thresholds. Each category contains 20 sentences generated using ChatGPT with the prompt "Provide 10 safe and 10 unsafe sentences related to Offshore Wind Turbine Maintenance within category $N$".
\\Figure \ref{fig_roc_curves} shows Receiver Operating Characteristic (ROC) curve of the classification performance of safe sentences for each of the 10 categories; \textit{(a)} using EMD, \textit{(b)} using Cosine similarity. Figure \ref{fig_roc_curves}(a) has Area Under the Curve (AUC) ranges from 0.65 (Emergency Procedures) to 0.98 (Risk Assessment), with a mean average of 0.78 across all categories. Figure \ref{fig_roc_curves}(b) has AUC ranges from 0.4 (Emergency Procedures) to 0.98 (Regulatory Compliance), with a mean average of 0.725.

\section{Future Work}
Through further development, it is hoped that industry engagement can be established with a fully functional demo version of SafeLLM, gaining valuable data to train and test the models. Further, a comprehensive dictionary of unsafe concepts with less generalised categories can be defined in line with current industry standards.

SafeLLM focuses on the functional safety of LLMs with less focus on the user Graphical User Interface (GUI). However, it is proposed in future work that the implementation of a GUI will benefit the project further and allow successful integration into industry. Future development of the conversation agent is hoped to introduce the framework as a training aid to maintenance personnel, with a natural language interface to break down complex tasks.

\begin{figure}[!ht]
\centering
\includegraphics[width=0.6\textwidth]{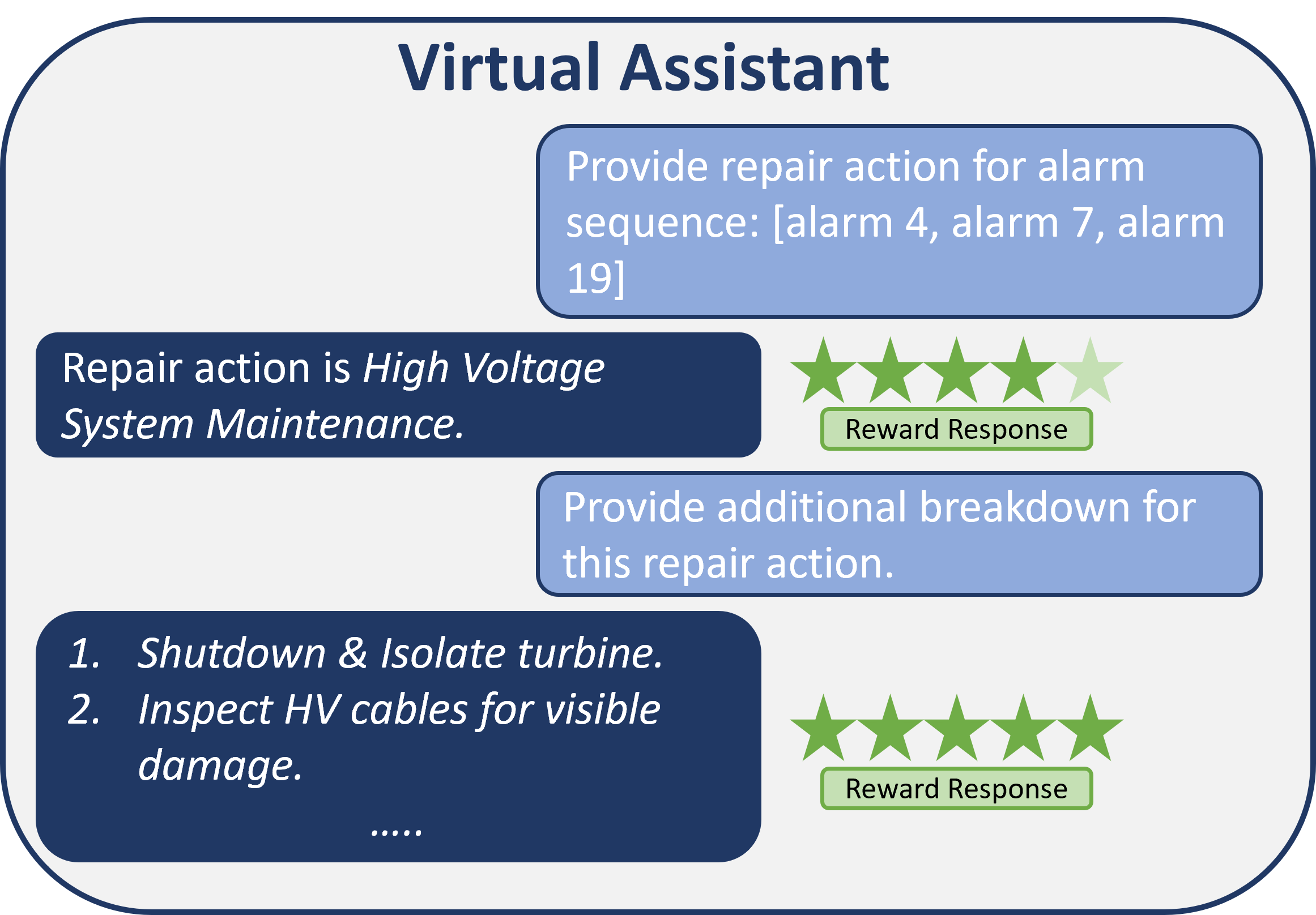}
\caption{\label{chatbot}An example breakdown of instructions provided to maintenance personnel.}    
\end{figure}

Figure \ref{chatbot} therefore suggests an example GUI for the O\&M manager to provide feedback on responses. This is designed to incorporate all discussed SafeLLM features, providing only suitable and safe responses to the maintenance personnel working on the turbines.
\section{Conclusion}
The paper has focused on the development of a specialised conversational agent that uses an input of alarm sequences to recommend repair actions. The evaluation has been limited in scope using ChatGPT-4 data. Results are therefore preliminary showing proof-of-concept and only potential success of EMD as a foundation for SafeLLM framework.
Overall, it has been shown that Cosine Similarity outperforms EMD in most of the categories we tested, but there is some sign of EMD having potential. We are currently re-training with larger, more informative OSW datasets in an industrial case study with Électricité de France, and we hope to throw more light on the efficacy of the approach in our next publication.

Further development of this approach is expected to contribute to reliable automation of maintenance tasks as well as training and assisting personnel in complex maintenance with associated benefits of improved efficiency and reduced costs. For example, by adding an additional layer to the current LLM, the framework may also benefit use in training personnel both before and on-site with repair action prompts breaking down larger tasks. 

Overall, our methodology represents a potential shift in maintenance practices, automating part of the task by leveraging LLMs to extract actionable maintenance insights from previously obscure data. 
\section*{Acknowledgement}
{ This work was conducted under the Aura CDT program, funded by EPSRC and NERC, grant number EP/S023763/1 \textsuperscript{1}, EP/S023763/1 \textsuperscript{2} and project reference 2609857 \textsuperscript{1}, 2609795 \textsuperscript{2}. This work was also supported by the Secure and Safe Multi-Robot Systems (SESAME) H2020 Project under Grant Agreement 101017258.
 \textsuperscript{1} Connor Walker, \textsuperscript{2} Callum Rothon}

\bibliography{References.bib}
\bibliographystyle{IEEEtran}

\end{document}